\documentclass[runningheads,orivec]{llncs}

\usepackage[T1]{fontenc}

\usepackage{graphicx}

\usepackage{microtype}
\usepackage{booktabs}
\usepackage{multirow}
\usepackage{amsmath}
\usepackage{arydshln}
\usepackage{adjustbox}
\usepackage{todonotes}
\usepackage{color,xcolor,colortbl}
\usepackage[framemethod=TikZ]{mdframed}
\usepackage{pgfplots}
\usepgfplotslibrary{fillbetween}
\usetikzlibrary{patterns}
\usepackage{subcaption}

\pgfplotsset{compat=1.18}

\newcommand{\quotes}[1]{``#1''}

\makeatletter
\def\adl@drawiv#1#2#3{%
        \hskip.5\tabcolsep
        \xleaders#3{#2.5\@tempdimb #1{1}#2.5\@tempdimb}%
                #2\z@ plus1fil minus1fil\relax
        \hskip.5\tabcolsep}
\newcommand{\cdashlinelr}[1]{%
  \noalign{\vskip\aboverulesep
           \global\let\@dashdrawstore\adl@draw
           \global\let\adl@draw\adl@drawiv}
  \cdashline{#1}
  \noalign{\global\let\adl@draw\@dashdrawstore
           \vskip\belowrulesep}}
\makeatother

\definecolor{lightblue}{HTML}{5DA5DA}
\definecolor{lightorange}{HTML}{FAA43A}
\definecolor{lightgreen}{HTML}{60BD68}
\definecolor{lightpurple}{HTML}{B276B2}
\definecolor{brown}{HTML}{D95F0E}
\definecolor{lightgrey}{HTML}{BDBDBD}

\begin{document}
\title{Large Language Models for Czech Aspect-Based Sentiment Analysis}

\author{Jakub \v{S}m\'{i}d\inst{1,2}\orcidID{0000-0002-4492-5481} \and Pavel P\v{r}ib\'{a}\v{n}\inst{1}\orcidID{0000-0002-8744-8726} \and Pavel Kr\'{a}l\inst{1,2}\orcidID{0000-0002-3096-675X}}
 \institute{University of West Bohemia in Pilsen\\
          Faculty of Applied Sciences, Department of Computer Science and Engineering\\
          \and
          NTIS -- New Technologies for the Information Society\\
          Univerzitni 27328, 301 00 Plze\v{n}, Czech Republic\\
          \email{\{jaksmid,pribanp,pkral\}@kiv.zcu.cz} \\
            \tt {\url{https://nlp.kiv.zcu.cz}} \\
          }

\maketitle
\begin{abstract}
Aspect-based sentiment analysis (ABSA) is a fine-grained sentiment analysis task that aims to identify sentiment toward specific aspects of an entity. While large language models (LLMs) have shown strong performance in various natural language processing (NLP) tasks, their capabilities for Czech ABSA remain largely unexplored. In this work, we conduct a comprehensive evaluation of 19 LLMs of varying sizes and architectures on Czech ABSA, comparing their performance in zero-shot, few-shot, and fine-tuning scenarios. Our results show that small domain-specific models fine-tuned for ABSA outperform general-purpose LLMs in zero-shot and few-shot settings, while fine-tuned LLMs achieve state-of-the-art results. We analyze how factors such as multilingualism, model size, and recency influence performance and present an error analysis highlighting key challenges, particularly in aspect term prediction. Our findings provide insights into the suitability of LLMs for Czech ABSA and offer guidance for future research in this area.
\keywords{Aspect-based sentiment analysis \and Sentiment analysis \and Large language models \and Prompting}
\end{abstract}

\section{Introduction}

Aspect-based sentiment analysis (ABSA) is a natural language processing (NLP) task extends traditional sentiment analysis by targeting specific entities and their aspects, determining sentiment for each rather than providing an overall polarity. ABSA involves three sentiment elements~\cite{SMID2025103073}: the aspect term~($a$), denoting the opinion target; the aspect category~($c$), representing an attribute of an entity; and the sentiment polarity~($p$), reflecting the emotional tone. For instance, in the sentence \textit{\quotes{Excellent soup}}, these elements correspond to \textit{\quotes{soup}}, \textit{\quotes{food quality}}, and \textit{\quotes{positive}}. Aspect terms may also be implicit, as in \textit{\quotes{Tasty!}}.

ABSA tasks vary in complexity depending on which elements they cover. Simple tasks, such as aspect term detection, focus on a single element. Recently more popular compound tasks integrate multiple sentiment elements, such as aspect category sentiment analysis (ACSA)~\cite{schmitt-etal-2018-joint}, end-to-end ABSA (E2E-ABSA)~\cite{wang2018towards}, aspect category term extraction (ACTE)~\cite{pontiki-etal-2015-semeval}, and target-aspect-term-detection (TASD)~\cite{tasd}. Table~\ref{tab:absa-tasks} shows the input and output format of selected ABSA tasks.

\begin{table}[ht!]
    \centering
     \caption{Outputs of selected ABSA tasks for input: \textit{\quotes{Tasty tea but rude staff}}.}
    \begin{adjustbox}{width=0.7\linewidth}
        \begin{tabular}{@{}lll@{}}
            \toprule
            \textbf{Task} &  \textbf{Output}     & \textbf{Example output}  \\                        \midrule
            E2E-ABSA      &  \{($a$, $p$)\}      & \{(\quotes{tea}, POS), (\quotes{staff}, NEG)\}               \\
            ACSA      &  \{($c$, $p$)\}      & \{(drinks, POS), (service, NEG)\}               \\
            ACTE          &  \{($a$, $c$)\}      & \{(\quotes{tea}, drinks), (\quotes{staff}, service)\}           \\
            TASD          & \{($a$, $c$, $p$)\} & \{(\quotes{tea}, drinks, POS), (\quotes{staff}, service, NEG)\} \\ \bottomrule
        \end{tabular}
    \end{adjustbox}
	\label{tab:absa-tasks}
\end{table}

While ABSA has been widely studied for English, other languages, including Czech, remain underrepresented. Early Czech ABSA studies~\cite{hercig2016unsupervised,steinberger-etal-2014-aspect} relied on now-outdated sentiment classification methods. Recent research~\cite{priban-prazak-2023-improving,smid-priban-2023-prompt,smid-etal-2024-czech} has adopted modern Transformer-based~\cite{NIPS2017_3f5ee243} approaches.

Large language models (LLMs), such as GPT-4o~\cite{openai2024gpt4ocard}, have transformed NLP via \textit{prompting}, a technique that replaces fine-tuning by guiding model behaviour through task instructions. Few-shot prompting -- providing input–output examples -- can further enhance performance. However, for complex tasks like ABSA, fine-tuned smaller models still tend to outperform general-purpose LLMs~\cite{bai-etal-2024-compound,zhang-etal-2024-sentiment}. Although LLM fine-tuning is resource-intensive, methods like QLoRA~\cite{qlora} reduce memory usage, enabling efficient fine-tuning on consumer GPUs. Fine-tuned LLMs using QLoRA have outperformed smaller models for ABSA~\cite{smid-etal-2024-llama}.

Despite these advancements, LLM-based ABSA for languages other than English is still underexplored~\cite{SMID2025103073}. Few studies have assessed LLM performance on multilingual ABSA tasks~\cite{icaart25,wu2024evaluatingzeroshotmultilingualaspectbased}, with no comprehensive evaluation for Czech ABSA. This study addresses this gap by evaluating multiple LLMs across zero-shot, few-shot, and fine-tuning scenarios on four Czech ABSA tasks.

Our main contributions include: 1) We provide a comprehensive evaluation of 19 large language models of varying architectures and sizes for Czech aspect-based sentiment analysis, being the first to do so. 2) We compare the zero-shot, few-shot, and fine-tuned performance of LLMs, showing that fine-tuned LLMs achieve new state-of-the-art results, while smaller ABSA-specific models from previous work outperform general-purpose LLMs in zero-shot and few-shot settings. 3) We provide an analysis of the impact of model properties, such as multilingualism, size, and recency, on ABSA performance. 4) We conduct a detailed error analysis identifying key challenges in Czech ABSA, particularly in aspect term prediction.

\section{Related Work}
Early Czech ABSA research~\cite{hercig2016unsupervised,steinberger-etal-2014-aspect} rely on traditional methods like conditional random fields and maximum entropy classifiers. Recent approaches adopt Transformer-based models. Some enhance ABSA with semantic role labelling in a multitask setup~\cite{priban-prazak-2023-improving}, while others explore prompt-based learning and the use of Czech-specific models and in-domain pre-training~\cite{smid-priban-2023-prompt,smid-etal-2024-czech}.

LLMs have been evaluated for ABSA, but fine-tuned smaller models often outperform LLMs in zero- and few-shot settings~\cite{bai-etal-2024-compound,zhang-etal-2024-sentiment}. Fine-tuning LLMs has been shown to improve performance across languages~\cite{smid-etal-2024-llama,icaart25,wu2024evaluatingzeroshotmultilingualaspectbased}, highlighting the value of task-specific fine-tuning.

\section{Experimental Setup}
We conduct experiments on ACSA, E2E-ABSA, ACTE, and TASD. We utilize the \texttt{CsRest-M} dataset~\cite{smid-etal-2024-czech} consisting of real-world restaurant reviews in Czech designed for compound ABSA tasks, with annotations linking aspect terms, aspect categories, and sentiment polarities. The dataset is already split into training, validation, and test sets. Table~\ref{tab:data_stats} shows the statistics of the dataset.

\begin{table}[ht!]
    \centering
    \caption{Statistics of the dataset.}
    % \begin{adjustbox}{width=0.5\linewidth}
    \begin{tabular}{@{}lrrr@{}}
        \toprule
        \textbf{Count} & \textbf{Train} & \textbf{Dev} & \textbf{Test} \\ \midrule
        Sentences   & 2,151 & 240   & 798 \\
        Triplets    & 4,386 & 483   & 1,609 \\ \bottomrule
    \end{tabular}
    % \end{adjustbox}

    \label{tab:data_stats}
\end{table}

\subsection{Models}
We utilize two closed-source LLMs and several open-source LLMs of varying sizes. Table~\ref{tab:models} provides an overview of the models used in this paper, including their sizes and language support. \textit{English-centric} indicates that while the models were primarily pre-trained and instruction-tuned in English, they may also include data from other languages\footnote{For example, approximately 90\% of LLaMA~2’s pre-training data is English~\cite{touvron2023llama2openfoundation}, with the remainder in other languages.}.

\begin{table}[ht!]
    \centering
    \caption{Alphabetically sorted LLMs used in our experiments, their sizes (in billions of parameters), and language support. $^\dagger$ indicates models with official support for Czech. * indicates models without official documentation on language support, assumed to be primarily English-centric.}
    \begin{adjustbox}{width=0.8\linewidth}
    \begin{tabular}{@{}llll@{}}
        \toprule
        \textbf{Model} & \textbf{Sizes (B)} & \textbf{Language Support} & \textbf{Open-source} \\ \midrule
        Aya~23 {\footnotesize \cite{aryabumi2024aya23openweight}}      & 8, 35 & Multilingual$^\dagger$ & Yes \\
        Gemma~3 {\footnotesize \cite{gemmateam2025gemma3technicalreport}}     & 1, 4, 12, 27 & 1B: English-centric, others: Multilingual$^\dagger$ & Yes \\
        GPT-3.5 Turbo  {\footnotesize \cite{openaigpt35turbo}}      & \multicolumn{1}{c}{--}            & Multilingual$^\dagger$ & No \\
        GPT-4o mini  {\footnotesize \cite{openai2024gpt4ocard}}      & \multicolumn{1}{c}{--}            & Multilingual$^\dagger$ & No \\
        LLaMA~2 {\footnotesize \cite{touvron2023llama2openfoundation}}        & 7, 13        & English-centric & Yes \\
        LLaMA~3 {\footnotesize \cite{dubey2024llama3herdmodels}}        & 8        & English-centric & Yes \\
        LLaMA~3.1 {\footnotesize \cite{dubey2024llama3herdmodels}}        & 8, 70        & Multilingual & Yes \\
        LLaMA~3.2 {\footnotesize \cite{dubey2024llama3herdmodels}}        & 1, 3        & Multilingual & Yes \\
        LLaMA~3.3 {\footnotesize \cite{dubey2024llama3herdmodels}}        & 70        & Multilingual & Yes \\
        Mistral (v0.3) {\footnotesize \cite{jiang2023mistral7b}}     & 7            & English-centric* & Yes \\
        Orca~2 {\footnotesize \cite{mitra2023orca2teachingsmall}}      & 7, 13        & English-centric* & Yes \\
        \bottomrule
    \end{tabular}
    \end{adjustbox}
    \label{tab:models}
\end{table}

\subsection{Prompting Strategy \& Fine-Tuning}
We design our prompts based on prior work~\cite{smid-etal-2024-llama,icaart25}, ensuring they are simple, clear, and standardized for ABSA. These prompts define sentiment elements and output format. Sentiment elements specify the permitted label space, such as aspect categories and sentiment polarities or that aspect terms must be found in the text or be \textit{\quotes{null}} for implicit ones, while the output format ensures consistency in model responses. We use the standard zero-shot prompt, as those have been shown to often outperform more complex strategies like chain-of-thought for E2E-ABSA in different languages~\cite{wu2024evaluatingzeroshotmultilingualaspectbased}.

\begin{figure}[ht!]
    \centering
    \includegraphics[width=0.99\linewidth]{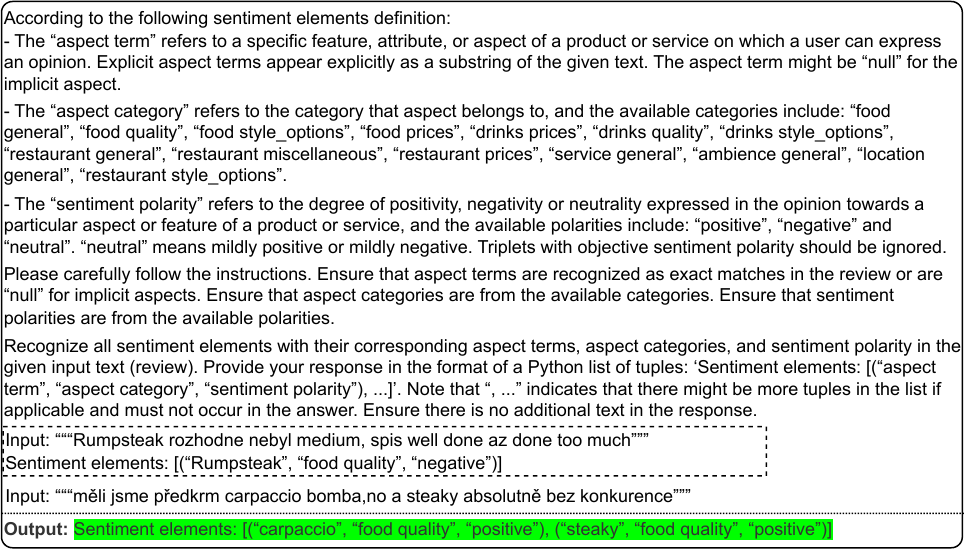}
    \caption{Prompt for the TASD task, showing an example input (English translation: \textit{\quotes{we had carpaccio as a starter -- amazing -- and the steaks were absolutely unmatched}}), the expected output in the green box, and one demonstration in the dashed box. Demonstrations are included only in few-shot scenarios.}
    \label{fig:prompt}
\end{figure}

Figure~\ref{fig:prompt} shows a TASD prompt, which we adapt for other tasks by omitting irrelevant elements (e.g. sentiment polarity for ACTE). For few-shot experiments, we use the first ten training examples due to their balanced label distribution.

We also test Czech-translated prompts, as prior work shows language alignment helps, especially with English-centric LLMs~\cite{liu2024translationneedstudysolving}. Instead of translating the dataset into English -- which risks misalignment and errors -- we translate the prompt to Czech to preserve evaluation quality.

For fine-tuning, we use QLoRA~\cite{qlora} on models up to 13B parameters, which adds LoRA~\cite{hu2022lora} weights to a 4-bit quantized backbone, reducing memory use while maintaining performance. Since prompt language has no effect during fine-tuning, we use English-only prompts and the task-specific training set, fine-tuning the model to generate outputs in the desired format.

\subsection{Experimental Details}

We use the official API\footnote{\url{https://platform.openai.com/docs/overview}} for GPT models but exclude GPT-3.5 Turbo with Czech prompts due to budget limits. For open-source LLMs, we use instruction-tuned models from HuggingFace Transformers~\cite{wolf-etal-2020-transformers}. We use 4-bit quantized models, which offer performance similar to 8-bit or full-precision versions~\cite{qlora}.

Fine-tuning follows QLoRA~\cite{qlora}, with 4-bit NF4 quantization, bf16 precision, AdamW~\cite{adamw}, a learning rate of 2e-4, batch size 16, and LoRA adapters on all linear layers. While $r=64$, $\alpha=16$ works for most models, Gemma~3 (4B/12B) required tuning. A grid search found $r=64$, $\alpha=128$ performed best. Models are trained for up to 5 epochs, selecting the best by validation loss. Following prior work~\cite{mitra2023orca2teachingsmall,smid-etal-2024-llama,icaart25}, we compute loss only on generated tokens. All experiments use greedy decoding and run on an NVIDIA L40 GPU with 48~GB of VRAM.

\subsection{Evaluation Metrics \& Compared Methods}
We use micro F1-score, a standard metric in ABSA research, and consider a predicted sentiment tuple correct only if all its elements match the gold tuple exactly. For fine-tuning experiments, we report the average results over five runs.

We compare the performance of LLMs against the best results reported in~\cite{smid-etal-2024-czech}, who fine-tuned multilingual and Czech-only Transformer-based models. For the ACSA task, there are no prior results on the employed dataset.

\section{Results}

\begin{table}[ht!]
\centering
\caption{Zero- and few-shot results on different tasks with English (En) and Czech (Cs) prompts with different LLMs compared to the best results with fine-tuned models achieved in~\cite{smid-etal-2024-czech}. For each column, the best result is in \textbf{bold}, the second best is \underline{underlined}. We group the LLMs by architecture and sort by size.}
\begin{adjustbox}{width=\linewidth}

\begin{tabular}{@{}lrrrr@{\hspace{8pt}}rrrr@{\hspace{8pt}}rrrr@{\hspace{8pt}}rrrr@{}}
\toprule
              & \multicolumn{4}{c}{\textbf{ACSA}}                                          & \multicolumn{4}{c}{\textbf{ACTE}}                                          & \multicolumn{4}{c}{\textbf{E2E-ABSA}}                                      & \multicolumn{4}{c}{\textbf{TASD}}                                          \\ \cmidrule(lr){2-5} \cmidrule(lr){6-9} \cmidrule(lr){10-13} \cmidrule(lr){14-17}
              & \multicolumn{2}{c}{Zero-shot}   & \multicolumn{2}{c}{Few-shot}    & \multicolumn{2}{c}{Zero-shot}   & \multicolumn{2}{c}{Few-shot}    & \multicolumn{2}{c}{Zero-shot}   & \multicolumn{2}{c}{Few-shot}    & \multicolumn{2}{c}{Zero-shot}   & \multicolumn{2}{c}{Few-shot}    \\
              \cmidrule(lr){2-3} \cmidrule(lr){4-5} \cmidrule(lr){6-7} \cmidrule(lr){8-9} \cmidrule(lr){10-11} \cmidrule(lr){12-13} \cmidrule(lr){14-15} \cmidrule(lr){16-17}
              & En             & Cs             & En             & Cs             & En             & Cs             & En             & Cs             & En             & Cs             & En             & Cs             & En             & Cs             & En             & Cs             \\ \midrule
\cite{smid-etal-2024-czech}              & \multicolumn{4}{c}{--}                                            & \multicolumn{4}{c}{67.30}                                         & \multicolumn{4}{c}{74.80}                                         & \multicolumn{4}{c}{59.30}                                         \\ \cdashlinelr{1-17}
GPT-3.5 Turbo & 57.29          & \multicolumn{1}{c}{--}             & 61.64          & \multicolumn{1}{c}{--}             & 26.32          & \multicolumn{1}{c}{--}             & 45.79          & \multicolumn{1}{c}{--}             & 44.58          & \multicolumn{1}{c}{--}             & 54.75          & \multicolumn{1}{c}{--}             & 25.39          & \multicolumn{1}{c}{--}             & 42.60          & \multicolumn{1}{c}{--}             \\
GPT-4o mini   & 61.43          & 61.65          & 69.90          & \underline{70.94}    & 34.22          & 21.30           & 51.75          & 49.32          & \textbf{54.38} & \underline{46.45}    & \underline{60.72}    & 59.51          & 35.53          & 24.18          & 46.07          & 46.21          \\ \cdashlinelr{1-17}
Aya 23 8B        & 41.86          & 43.26          & 61.81          & 62.53          & 17.70          & 9.38           & 39.60          & 38.33          & 26.16          & 16.50          & 47.66          & 44.50          & 13.74          & 6.99           & 35.62          & 35.67          \\
Aya 23 35B       & 61.67          & 61.75          & 67.00          & 67.27          & 28.43          & 26.94          & 52.88          & 53.15          & 43.94          & 28.90          & 59.79          & 55.80          & 25.98          & 25.71          & 46.34          & 48.37          \\ \cdashlinelr{1-17}
Gemma 3 1B    & 5.52           & 2.64           & 38.20          & 32.01          & 4.99           & 0.55           & 19.12          & 13.91          & 4.39           & 0.08           & 23.87          & 15.95          & 5.39           & 0.72           & 14.96          & 12.55          \\
Gemma 3 4B    & 57.82          & 59.66          & 63.13          & 65.12          & 39.34          & 18.26          & 49.21          & 48.86          & 42.43          & 32.18          & 54.35          & 52.46          & 32.68          & 13.97          & 47.72          & 44.56          \\
Gemma 3 12B   & \underline{69.25}    & \underline{69.93}    & \underline{69.97}    & 69.27          & \underline{49.24}    & \underline{41.24}    & \underline{56.65}    & \underline{56.79}    & \underline{53.98}    & 45.85          & 59.81          & \underline{59.81}    & \underline{44.61}    & \underline{37.10}    & \underline{51.66}    & \underline{52.47}    \\
Gemma 3 27B   & \textbf{69.79} & \textbf{70.91} & \textbf{72.76} & \textbf{72.74} & \textbf{51.47} & \textbf{47.18} & \textbf{58.60} & \textbf{58.26} & 51.89          & \textbf{47.44} & \textbf{64.23} & \textbf{63.65} & \textbf{46.68} & \textbf{41.89} & \textbf{54.53} & \textbf{54.64} \\ \cdashlinelr{1-17}
LLaMA 2 7B    & 14.15          & 3.82           & 40.96          & 40.47          & 5.97           & 0.46           & 29.08          & 32.61          & 12.24          & 0.74           & 35.04          & 37.94          & 3.58           & 0.94           & 25.94          & 27.06          \\
LLaMA 2 13B   & 32.73          & 27.78          & 49.03          & 52.17          & 9.57           & 4.94           & 37.66          & 35.86          & 18.95          & 13.21          & 44.03          & 44.03          & 10.21          & 6.43           & 35.73          & 35.72          \\ \cdashlinelr{1-17}
LLaMA 3 8B    & 53.32          & 3.01           & 58.97          & 47.28          & 16.74          & 2.80           & 39.45          & 31.64          & 34.54          & 11.79          & 42.32          & 39.18          & 7.91           & 8.31           & 34.64          & 28.86          \\ \cdashlinelr{1-17}
LLaMA 3.1 8B  & 29.72          & 26.95          & 48.28          & 1.92           & 8.90           & 12.24          & 27.36          & 7.62           & 12.30          & 23.19          & 41.65          & 6.22           & 11.51          & 1.97           & 22.31          & 0.12           \\
LLaMA 3.1 70B & 55.15          & 54.53          & 68.58          & 67.47          & 27.04          & 24.99          & 50.62          & 51.13          & 44.37          & 37.84          & 59.38          & 57.35          & 26.08          & 23.33          & 47.79          & 45.47          \\ \cdashlinelr{1-17}
LLaMA 3.2 1B  & 0.12           & 2.76           & 0.12           & 1.09           & 0.00           & 0.11           & 0.85           & 0.00           & 0.00           & 0.22           & 0.00           & 0.00           & 0.00           & 0.00           & 0.00           & 0.00           \\
LLaMA 3.2 3B  & 0.00           & 6.96           & 0.00           & 2.55           & 0.89           & 0.47           & 2.40           & 2.02           & 0.12           & 3.91           & 9.37           & 1.06           & 0.00           & 0.71           & 3.59           & 0.00           \\ \cdashlinelr{1-17}
LLaMA 3.3 70B & 55.59          & 54.41          & 70.08          & 68.75          & 28.35          & 25.18          & 52.92          & 53.54          & 48.89          & 42.46          & 59.20          & 54.15          & 27.85          & 24.20          & 49.72          & 47.92          \\ \cdashlinelr{1-17}
Mistral 7B    & 43.56          & 47.32          & 57.17          & 56.12          & 11.63          & 7.55           & 41.13          & 37.83          & 21.47          & 17.21          & 44.52          & 39.24          & 11.58          & 8.76           & 37.22          & 32.63          \\ \cdashlinelr{1-17}
Orca 2 7B     & 35.73          & 0.95           & 54.28          & 53.29          & 7.61           & 1.23           & 28.43          & 27.54          & 16.06          & 6.19           & 32.97          & 31.16          & 4.58           & 0.43           & 26.75          & 17.68          \\
Orca 2 13B    & 49.51          & 45.72          & 63.39          & 62.88          & 13.72          & 11.49          & 35.09          & 34.81          & 22.67          & 20.81          & 41.04          & 39.99          & 11.19          & 11.19          & 32.63          & 32.66          \\ \bottomrule
\end{tabular}
\end{adjustbox}
\label{tab:res}
\end{table}

Table~\ref{tab:res} presents the zero-shot and few-shot results with Czech and English prompts on four ABSA tasks with different LLMs compared to fine-tuned models. There are several observations:
\\
\hspace*{5pt}
1) \textbf{Effect of Prompt Language}: The impact of using Czech versus English prompts is inconsistent. While Czech prompts sometimes yield slightly better results, English prompts generally perform better. In some cases, the differences are significant; for instance, in the zero-shot ACSA task, LLaMA~3~8B performs about 50\% better with an English prompt than a Czech one. However, such large margins are uncommon.
\\
\hspace*{5pt}
2) \textbf{Impact of Model Size, Recency, and Multilingualism}: As expected, larger, newer, and multilingual models tend to achieve better results. Older models such as Orca~2 and LLaMA~2 significantly underperform compared to more recent multilingual models of similar or even smaller sizes. Additionally, despite being multilingual, LLaMA~3.2 models perform extremely poorly, often scoring 0\%. Upon closer examination, we found that these models generated Python code instead of task-relevant outputs, suggesting they failed to understand the task. Interestingly, even few-shot prompting does not help these models. Similarly, Gemma~3~1B struggles in zero-shot scenarios but improves substantially when provided with few-shot examples.
\\
\hspace*{5pt}
3) \textbf{Effect of Few-Shot Examples}: Providing few-shot examples generally improves results, particularly for smaller and more English-centric models. These findings suggest that these models struggle to understand the task from a zero-shot prompt alone, but demonstrations help guide them toward the correct interpretation.
\\
\hspace*{5pt}
4) \textbf{Performance of Proprietary Models}: Among proprietary models, GPT-4o mini consistently outperforms GPT-3.5 Turbo, likely due to its newer architecture and improved capabilities. 
\\
\hspace*{5pt}
5) \textbf{Strong Performance of Aya and Gemma Models}: Among open-source models, the Aya and Gemma models perform particularly well, likely due to their official support for Czech and recent release. Notably, Gemma~3~27B performs best in most cases, with Gemma~3~12B frequently ranking second. Their strong results are particularly impressive given that they often outperform proprietary GPT models with significantly more parameters. Aya~23~35B is usually about 5\% worse than the Gemma~3~27B model and only slightly worse than Gemma~3~12B in few-shot scenarios. However, the difference in zero-shot settings is larger; for example, Aya~23~35B is about 20\% worse for TASD than Gemma~3~27B. The smaller 8B version of Aya~23 is often more than 10\% worse than the 35B version, while the 4B version of Gemma~3 is about 10\% worse than the 12B version and is often comparable or only slightly worse than the much larger Aya~23~35B.
\\
\hspace*{5pt}
6) \textbf{Task Difficulty Ranking}: The models generally perform best on ACSA, followed by E2E-ABSA and ACTE, with TASD being the most challenging task. This ranking likely reflects differences in label complexity. ACSA is the easiest because it does not require predicting aspect terms, whereas ACTE and E2E-ABSA involve more complex label spaces. TASD is the hardest since it requires predicting three sentiment elements rather than just two.
\\
\hspace*{5pt}
7) \textbf{Comparison to Fine-Tuned Models}: The best-performing LLMs achieve zero-shot results approximately 20\% lower than fine-tuned models. With few-shot prompting, this gap shrinks to around 5--10\%. While fine-tuned models still offer superior performance, LLMs provide a viable alternative when annotated data is scarce. Their ability to generate results quickly without the need for fine-tuning makes them attractive for rapid deployment, though fine-tuned models remain the preferred choice when performance is the primary concern.

\begin{table}[ht!]
\centering
\caption{Results with different fine-tuned LLMs compared to the best results with fine-tuned models achieved in~\cite{smid-etal-2024-czech}, alongside the average score. For each task, the best result is in \textbf{bold}, the second best is \underline{underlined}.}
\begin{adjustbox}{width=0.52\linewidth}
\begin{tabular}{@{}lr@{\hspace{6pt}}r@{\hspace{6pt}}r@{\hspace{6pt}}r@{\hspace{6pt}}r@{}}
\toprule
             & \textbf{ACSA}          & \textbf{ACTE}  & \textbf{E2E} & \textbf{TASD}  & \textbf{AVG}   \\ \midrule
\cite{smid-etal-2024-czech}             & \multicolumn{1}{c}{--} & 67.30          & 74.80             & 59.30          &  \multicolumn{1}{c}{--}          \\ \cdashlinelr{1-6}
Aya 23 8B       & 76.62                  & 73.02          & 74.04             & 68.08          & 72.94          \\ \cdashlinelr{1-6}
Gemma 3 1B   & 68.09                  & 63.52          & 64.74             & 53.68          & 62.50          \\
Gemma 3 4B   & 73.00                  & 70.57          & 73.02             & 65.27          & 70.46          \\
Gemma 3 12B  & \underline{76.78}            & \underline{74.30}    & \textbf{75.10}    & \textbf{69.36} & \underline{73.89}    \\ \cdashlinelr{1-6}
LLaMA 2 7B   & 73.31                  & 66.13          & 66.53             & 60.20          & 66.54          \\
LLaMA 2 13B  & 73.17                  & 67.01          & 69.39             & 60.75          & 67.58          \\ \cdashlinelr{1-6}
LLaMA 3 8B   & 70.77                  & 63.07          & 62.97             & 56.84          & 63.41          \\ \cdashlinelr{1-6}
LLaMA 3.1 8B & \textbf{77.51}         & \textbf{75.46} & \textbf{75.10}    & \underline{69.06}    & \textbf{74.28} \\ \cdashlinelr{1-6}
LLaMA 3.2 1B & 65.26                  & 64.16          & 63.35             & 55.71          & 62.12          \\
LLaMA 3.2 3B & 73.75                  & 69.14          & 68.54             & 61.07          & 68.13          \\ \cdashlinelr{1-6}
Mistral 7B   & 61.13                  & 55.14          & 54.41             & 48.52          & 54.80          \\ \cdashlinelr{1-6}
Orca 2 7B    & 74.26                  & 69.99          & 70.75             & 63.36          & 69.59          \\
Orca 2 13B   & 75.37                  & 72.61          & 71.83             & 65.62          & 71.36          \\ \bottomrule
\end{tabular}
\end{adjustbox}
\label{tab:res_fine}
\end{table}

Table~\ref{tab:res_fine} presents the results with fine-tuned models, showing significant improvements over previous state-of-the-art approaches. The largest gain is observed in the TASD task, where our best-performing model surpasses prior results by approximately 10\%. The top-performing models are LLaMA~3.1~8B, Gemma~3~12B, and Aya~23~8B, demonstrating the effectiveness of fine-tuning for enhancing LLM-based sentiment analysis. Notably, fine-tuning yields greater improvements for English-centric models than multilingual ones, suggesting that language-specific adaptations play a crucial role. Mistral~7B achieves the lowest scores, possibly due to suboptimal training hyperparameters rather than inherent model limitations. The results with 1B and 3B models improve substantially over the zero-shot and few-shot performance, even by 70\% in some cases. These results confirm that fine-tuned LLMs are strong alternatives to traditional models for ABSA tasks not only in English, but also in Czech.

\subsection{Effect of Few-Shot Example Count}

\begin{figure}[ht!]
    \centering
    \begin{adjustbox}{width=0.48\linewidth}
        \begin{tikzpicture}
            \begin{axis}[
                xlabel={\small{Number of few-shot examples}},
                ylabel={\small{F1-score [\%]}},
                xmin=0, xmax=21,
                xtick=data,
                xticklabels={0, 1, 2, 5, 10, 15,  20},
                ymin=0, ymax=70,
                ytick={10,20,30,40,50,60},
                ymajorgrids=true,
                xmajorgrids=true,
                grid style=dashed,
                legend style={font=\footnotesize,at={(0.08,0.23)},anchor=west},
                ]

                \addplot[
                    color=orange,
                    mark=diamond, 
                    ultra thick,
                ] 
                coordinates {
                    (0,24.8647730797529)
                    (1,35.5133861303329)
                    (2,37.9630338892936)
                    (5,43.1173525750637)
                    (10,46.1730167269706)
                    (15,46.7910572886467)
                    (20,46.6353185474872)
                };
                \addlegendentry{Aya~23~8B}

                 \addplot[
                    color=green,
                    mark=star, 
                    ultra thick,
                ] 
                coordinates {
                    (0,40.005549788475)
                    (1,47.7136477828026)
                    (2,53.7301503121853)
                    (5,52.4554260075092)
                    (10,56.1444878578185)
                    (15,57.2758480906486)
                    (20,57.5605705380439)
                };
                \addlegendentry{Aya~23~35B}

                \addplot[
                    color=blue,
                    mark=triangle*,
                    ultra thick,
                ]
                coordinates {
                    (0,54.2695820331573)
                    (1,56.8431630730629)
                    (2,57.3486089706421)
                    (5,58.7082907557487)
                    (10,59.5240131020546)
                    (15,62.31223549366)
                    (20,58.7390154099464)
                };
                \addlegendentry{Gemma~3~12B}

                 \addplot[
                    color=red,
                    mark=*,
                    ultra thick,
                ]
                coordinates {
                    (0,54.9551382660865)
                    (1,59.8791480064392)
                    (2,60.3509098291397)
                    (5,62.4444469809532)
                    (10,62.5282600522041)
                    (15,63.413242418766)
                    (20,62.324224011898)
                };
                \addlegendentry{Gemma~3~27B}

                 \addplot[
                    color=black,
                    mark=halfdiamond*,
                    ultra thick,
                ]
                coordinates {
                    (0,15.6058588996529)
                    (1,27.5238275527953)
                    (2,34.7713709578514)
                    (5,34.7639143466949)
                    (10,34.8969634622335)
                    (15,26.8139049410819)
                    (20,24.0702092647552)
                };
                \addlegendentry{LLaMA~3.1~8B}

                 \addplot[
                    color=purple,
                    mark=pentagon*,
                    ultra thick,
                ]
                coordinates {
                    (0,40.1704974472522)
                    (1,44.7318814694881)
                    (2,52.6840463280677)
                    (5,55.0998739898205)
                    (10,57.979018241167)
                    (15,59.5145285129547)
                    (20,59.2925891280174)
                };
                \addlegendentry{LLaMA~3.3~70B}
            \end{axis}
        \end{tikzpicture}
	\end{adjustbox}
	\caption{Impact of the number of few-shot examples on model performance. Results are averaged across all four tasks.}
	\label{fig:few}
\end{figure}
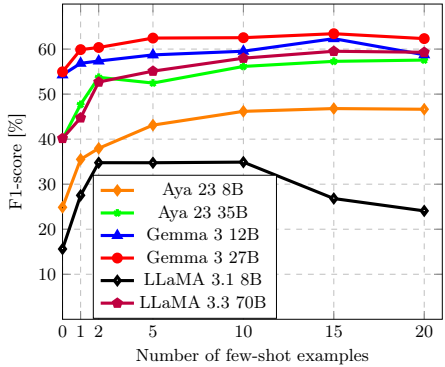

We analyze how the number of few-shot examples impacts performance for selected models. Figure~\ref{fig:few} presents the results, averaged across tasks, as their behaviour is generally consistent. Even a single few-shot example provides a noticeable improvement over zero-shot performance. Generally, increasing the number of examples leads to better results, though gains tend to plateau around 5 to 10 examples. Notably, LLaMA~3.1~8B exhibits a performance drop beyond 10 examples, primarily due to declines in ACSA and ACTE tasks. Given these trends, our choice of 10 few-shot examples appears to be a reasonable balance between performance gains and diminishing returns.

\subsection{Error Analysis}

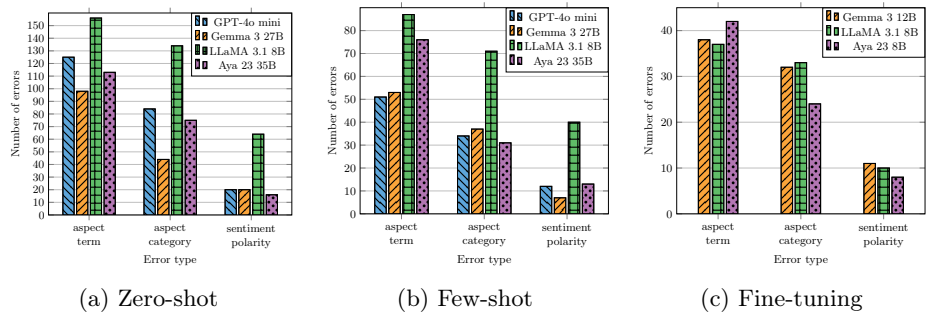
\begin{figure}[ht!]
    \begin{subfigure}{0.31\textwidth}
        \centering
        \begin{adjustbox}{width=\linewidth}
            \begin{tikzpicture}
                \begin{axis}[
                    ybar,
                    bar width=9pt,
                    xtick={0,1,2,3},
                    ytick={0,10,20,30,40,50,60,70,80,90,100,110,120,130,140,150},
                    xticklabel style={align=center,font=\small},
                    xticklabels={aspect\\term, aspect\\category, sentiment\\polarity},
                    ymin=0,
                    ymax=160,
                    xmin=-0.5,
                    xmax=2.5,
                    ymajorgrids=true,
                    ylabel={\small Number of errors},
                    xlabel={\small Error type},
                    legend style={at={(1,1)}, anchor=north east, font=\footnotesize},
                    ]
                    \addplot[black,fill=lightblue,postaction={pattern=north west lines}] coordinates {(0,125) (1,84) (2,20)};
                    \addplot[black,fill=lightorange,postaction={pattern=north east lines}] coordinates {(0,98) (1,44) (2,20)};
                    \addplot[black,fill=lightgreen,postaction={pattern=grid}] coordinates {(0,156) (1,134) (2,64)};
                    \addplot[black,fill=lightpurple,postaction={pattern=crosshatch dots}] coordinates {(0,113) (1,75) (2,16)};
                    \legend{GPT-4o mini, Gemma~3~27B,LLaMA~3.1~8B,Aya~23~35B}
                \end{axis}
            \end{tikzpicture}
        \end{adjustbox}
       \subcaption{Zero-shot}
    \end{subfigure}
    \hfill
    \begin{subfigure}{0.31\textwidth}
        \centering
        \begin{adjustbox}{width=\linewidth}
            \begin{tikzpicture}
                \begin{axis}[
                    ybar,
                    bar width=9pt,
                    xtick={0,1,2,3},
                    ytick={0,10,20,30,40,50,60,70,80},
                    xticklabel style={align=center,font=\small},
                    xticklabels={aspect\\term, aspect\\category, sentiment\\polarity},
                    ymin=0,
                    ymax=90,
                    xmin=-0.5,
                    xmax=2.5,
                    ymajorgrids=true,
                    ylabel={\small Number of errors},
                    xlabel={\small Error type},
                    legend style={at={(1,1)}, anchor=north east, font=\footnotesize},
                    ]
                    \addplot[black,fill=lightblue,postaction={pattern=north west lines}] coordinates {(0,51) (1,34) (2,12)};
                    \addplot[black,fill=lightorange,postaction={pattern=north east lines}] coordinates {(0,53) (1,37) (2,7)};
                    \addplot[black,fill=lightgreen,postaction={pattern=grid}] coordinates {(0,87) (1,71) (2,40)};
                    \addplot[black,fill=lightpurple,postaction={pattern=crosshatch dots}] coordinates {(0,76) (1,31) (2,13)};
                    \legend{GPT-4o mini, Gemma~3~27B,LLaMA~3.1~8B,Aya~23~35B}
                \end{axis}
            \end{tikzpicture}
        \end{adjustbox}
       \subcaption{Few-shot}
    \end{subfigure}
    \hfill
    \begin{subfigure}{0.31\textwidth}
        \centering
        \begin{adjustbox}{width=\linewidth}
            \begin{tikzpicture}
                \begin{axis}[
                    ybar,
                    bar width=9pt,
                    xtick={0,1,2,3},
                    ytick={0,10,20,30,40},
                    xticklabel style={align=center,font=\small},
                    xticklabels={aspect\\term, aspect\\category, sentiment\\polarity},
                    ymin=0,
                    ymax=45,
                    xmin=-0.5,
                    xmax=2.5,
                    ymajorgrids=true,
                    ylabel={\small Number of errors},
                    xlabel={\small Error type},
                    legend style={at={(1,1)}, anchor=north east, font=\footnotesize},
                    ]
                    \addplot[black,fill=lightorange,postaction={pattern=north east lines}] coordinates {(0,38) (1,32) (2,11)};
                    \addplot[black,fill=lightgreen,postaction={pattern=grid}] coordinates {(0,37) (1,33) (2,10)};
                    \addplot[black,fill=lightpurple,postaction={pattern=crosshatch dots}] coordinates {(0,42) (1,24) (2,8)};
                    \legend{Gemma~3~12B,LLaMA~3.1~8B,Aya~23~8B}
                \end{axis}
            \end{tikzpicture}
        \end{adjustbox}
       \subcaption{Fine-tuning}
    \end{subfigure}
    \caption{Error type distribution for different models on 100 TASD task examples.}
    \label{fig:error}
\end{figure}

We conduct an error analysis to evaluate model performance and identify key challenges. For this purpose, we randomly select 100 test examples and assess multiple models using the same set. Our analysis focuses on the TASD task in zero-shot, few-shot, and fine-tuning scenarios with an English prompt, manually comparing model predictions to ground truth labels. Figure~\ref{fig:error} presents the results.

Aspect term prediction poses the greatest challenge, as aspect terms can be any word or phrase in the text. Common errors include missing aspect terms, incorrect spans, and partial matches (e.g. omitting or adding words). Implicit aspect terms are particularly problematic – models frequently fail to recognize them or incorrectly predict explicit terms from the text instead. Notably, Aya~23~35B in zero-shot scenarios frequently predicts implicit aspect terms, though often incorrectly, whereas other models rarely identify implicit aspects at all. Additionally, in some cases, models predict aspect terms in their base (nominative) form, even when they appear in a different grammatical case in the text. For example, a model may predict \textit{\quotes{obsluha}} (\textit{\quotes{service}}) instead of the instrumental form \textit{\quotes{obsluhou}}. While technically a mismatch, such predictions are not necessarily incorrect. We recommend developing improved evaluation metrics tailored to LLMs, as the strict matching criteria commonly used in ABSA can be overly harsh in these situations and may unfairly penalize otherwise valid predictions.

Aspect category prediction is relatively easier due to the limited label space. However, models struggle with semantically similar categories, such as \textit{\quotes{restaurant miscellaneous}} and \textit{\quotes{restaurant general}}, and with rare categories like \textit{\quotes{location general}}. LLaMA~3.1~8B exhibits notably higher error rates in aspect category prediction in zero-shot and few-shot settings compared to other models.

Sentiment polarity prediction is the easiest task, with most errors occurring in the \textit{\quotes{neutral}} class. Models often misclassify mildly positive or mildly negative sentiment, which implies \textit{\quotes{neutral}} polarity, as \textit{\quotes{positive}} or \textit{\quotes{negative}}, respectively. These errors are significantly less frequent than those related to aspect terms or categories, likely because traditional sentiment analysis is well-represented in pre-training and instruction tuning data for LLMs.

Our analysis reveals that few-shot prompting reduces errors across all sentiment elements, with the greatest impact on aspect term prediction. Sentiment polarity, already the least error-prone element, benefits the least from it.

Fine-tuned models produce the fewest errors, particularly in aspect term prediction. Interestingly, all evaluated models in all scenarios incorrectly predicted the sentiment polarity for the phrase \textit{\quotes{Fajn bar}} (\textit{\quotes{Cool bar}}) as negative, while the correct sentiment polarity is \textit{\quotes{positive}}. The term \textit{\quotes{Fajn}} is from Common Czech, suggesting that the models struggle with these types of vernacular expressions.

\section{Conclusion}
This paper comprehensively evaluates large language models for Czech aspect-based sentiment analysis. We compare 19 LLMs of varying sizes and architectures, assessing their performance across zero-shot, few-shot, and fine-tuning scenarios. Our results highlight the strong influence of model properties -- such as multilingualism, size, and recency -- on ABSA performance. We find that small models fine-tuned specifically for ABSA outperform LLMs in zero-shot and few-shot settings, while fine-tuned LLMs achieve state-of-the-art results. Additionally, our error analysis identifies key challenges in Czech ABSA, offering insights into the strengths and limitations of LLMs for this task.

\section*{Acknowledgements}
This work has been supported by the Grant No. SGS-2025-022 -- New Data Processing Methods in Current Areas of Computer Science. Computational resources were provided by the e-INFRA CZ project (ID:90254), supported by the Ministry of Education, Youth and Sports of the Czech Republic.

\bibliographystyle{splncs04}
\bibliography{bibliography}

\end{document}